\documentclass[10pt,twocolumn,letterpaper]{article}

\usepackage{cvpr}              

\usepackage{float}

\definecolor{cvprblue}{rgb}{0.21,0.49,0.74}
\usepackage[pagebackref,breaklinks,colorlinks,allcolors=cvprblue]{hyperref}

\def\confName{CVPR}
\def\confYear{2026}

\title{PSR: Scaling Multi-Subject Personalized Image Generation with Pairwise Subject-Consistency Rewards}

\author{\textbf{Shulei Wang}$^{1}$ \quad \textbf{Longhui Wei}$^{2,\dagger}$ \quad \textbf{Xin He}$^{2}$\quad \textbf{Jianbo Ouyang}$^{2}$\quad \textbf{Hui Lu}$^{2}$ \\ \quad \textbf{Zhou Zhao}$^{1}$   \textbf{Qi Tian}$^{2}$
\\
$^{1}$ Zhejiang University\qquad
$^{2}$ Huawei Inc. \\
{\tt\small shuleiwang@zju.edu.cn, weilh2568@gmail.com}
}

\usepackage[misc]{ifsym}
\newcommand\blfootnote[1]{
    \begingroup
    \renewcommand\thefootnote{}\footnote{#1}
    \addtocounter{footnote}{-1}
    \endgroup
}
\usepackage[hang]{footmisc}

\begin{document}
\maketitle

{
    \blfootnote{ 
        $^\dagger$Corresponding author. 
        }
}

\begin{abstract}
Personalized generation models for a single subject have demonstrated remarkable effectiveness, highlighting their significant potential. However, when extended to multiple subjects, existing models often exhibit degraded performance, particularly in maintaining subject consistency and adhering to textual prompts. We attribute these limitations to the absence of high-quality multi-subject datasets and refined post-training strategies. To address these challenges, we propose a scalable multi-subject data generation pipeline that leverages powerful single-subject generation models to construct diverse and high-quality multi-subject training data. Through this dataset, we first enable single-subject personalization models to acquire knowledge of synthesizing multi-image and multi-subject scenarios. Furthermore, to enhance both subject consistency and text controllability, we design a set of Pairwise Subject-Consistency Rewards and general-purpose rewards, which are incorporated into a refined reinforcement learning stage. To comprehensively evaluate multi-subject personalization, we introduce a new benchmark that assesses model performance using seven subsets across three dimensions. Extensive experiments demonstrate the effectiveness of our approach in advancing multi-subject personalized image generation. Github Link: {\tt\small \href{https://github.com/wang-shulei/PSR}{https://github.com/wang-shulei/PSR}}.

\end{abstract}


\section{Introduction}
\label{sec:intro}
Personalized image generation aims to produce images that remain faithful to the given subjects~\cite{photomaker,instantid} while following textual instructions~\cite{wang2025towards,E240151}, and has significant applications in film production, personalized marketing, and beyond. Single-subject personalization models, such as FLUX.1 Kontext~\citep{labs2025flux} and Qwen-Image-Edit~\cite{qwen-image}, have already demonstrated impressive capabilities. Meanwhile, several recent efforts, including UNO~\citep{wu2025less} and OmniGen~\citep{xiao2025omnigen,wu2025omnigen2}, have begun to explore the domain of multi-subject generation, enabling models to accept multiple reference images and roughly maintain the overall subject consistency. However, these multi-subject personalization approaches still suffer from several limitations: \textbf{(a) Poor subject consistency}: the subjects in the generated images may not be similar to the given reference subjects, or even omit certain subjects entirely. \textbf{(b) Limited adherence to text prompts}: given the prompt ``the dog wears a chef’s hat and the cat wears a scarf", existing methods may fail to capture specific attributes, such as a chef’s hat, or incorrectly assign them. For instance, the generated image may depict the dog wearing a scarf and the cat wearing a chef’s hat, failing to follow the semantics specified in the prompt.





We attribute these shortcomings to two factors: the lack of high-quality multi-subject personalization datasets and the absence of refined post-training strategies. For multi-subject personalization datasets, existing methods such as OmniGen~\citep{xiao2025omnigen} introduce X2I-subject-driven, where most of the high-quality data focuses on human faces, while subject datasets for general scenarios are constructed from GRIT~\citep{peng2023kosmos} and exhibit relatively low consistency. UNO~\citep{wu2025less}, on the other hand, generates subject pairs using a Text-to-Image (T2I) model, which inherently introduces discrepancies in consistency.

\begin{figure*}[ht]
\centering
\includegraphics[width=0.85\linewidth]{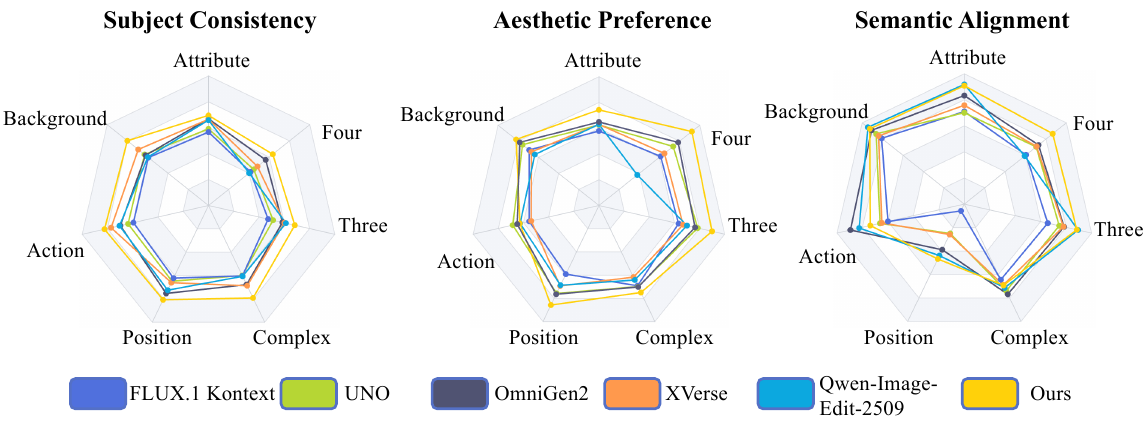}
\caption{Quantitative comparison of recent methods on PSRBench across three evaluation dimensions: subject consistency, aesthetic preference, and semantic alignment. Our method consistently outperforms all baselines across all seven subsets.}
\label{fig:quan}
\vspace{-0.4cm}
\end{figure*}
To address this, we leverage single-subject personalization models such as FLUX.1 Kontext~\citep{labs2025flux} to introduce a scalable multi-subject data generation pipeline. Specifically, we first employ T2I models~\cite{flux,qwen-image} to synthesize images containing multi-subject scenes. Then, a detection model~\cite{groundingDino} is applied to extract single-object instances, which are further used to generate input images through a single subject personalization model~\cite{labs2025flux}, thereby constructing multi-subject synthetic image pairs. This pipeline yields a large-scale dataset of 350K multi-subject images. With these data in place, we first conduct supervised fine-tuning(SFT) on a single-subject personalization model and introduce a scalable frame-wise positional encoding that equips the model with multi-subject personalization knowledge. This encoding scheme generalizes effectively to varying numbers of input reference images. Moreover, existing methods are limited to the SFT stage, where the optimization objective is defined at the global image level, making it difficult to ensure subject consistency. To address this limitation, we propose the Pairwise Subject-Consistency Rewards (PSR), which is combined with other generation rewards to improve subject consistency and text compliance.

Additionally, existing benchmarks for multi-subject personalized generation are neither fine-grained nor comprehensive, which hinders accurate evaluation of model performance. For instance, DreamBench~\cite{ruiz2023dreambooth} simply relies on DINO~\cite{dinov2} similarity between the generated image and the reference subject at the global level, while XVerseBench~\cite{chen2025xverse} evaluates generated images via detection and segmentation but does not segment the reference subject images, nor does it consider scenarios with more subjects. Therefore, to comprehensively evaluate multi-subject personalization, we propose PSRBench, a new benchmark with seven subsets, each assessing models from three dimensions: subject consistency, aesthetic preference, and semantic alignment.




Based on PSRBench, we evaluate our model against existing state-of-the-art approaches. As detailed in Figure~\ref{fig:quan} and  Sec.~\ref{sec:qual}, extensive experiments demonstrate that our method achieves SOTA performance across multiple subsets, with both quantitative and qualitative results validating its effectiveness.


Our contributions are summarized as follows:
\begin{itemize}
    \item We propose a scalable multi-subject personalization data generation pipeline that leverages existing single-subject models to synthesize data with arbitrary numbers of subjects. Using this pipeline, together with predefined data-cleaning strategies, we construct approximately 350K high-quality samples for multi-subject personalized generation.

    \item We extend the scalable frame-wise positional encoding and introduce Pairwise Subject-Consistency Rewards, which together, through a two-stage training paradigm, substantially enhance the model’s ability to maintain subject consistency and to faithfully adhere to textual instructions.

    \item We present PSRBench, a fine-grained and comprehensive benchmark with seven subsets, evaluating multi-subject personalization across three dimensions: subject consistency, aesthetic preference, and semantic alignment.
\end{itemize}

\section{Related Works}
\begin{figure*}[h]
\centering
\includegraphics[width=0.86\linewidth]{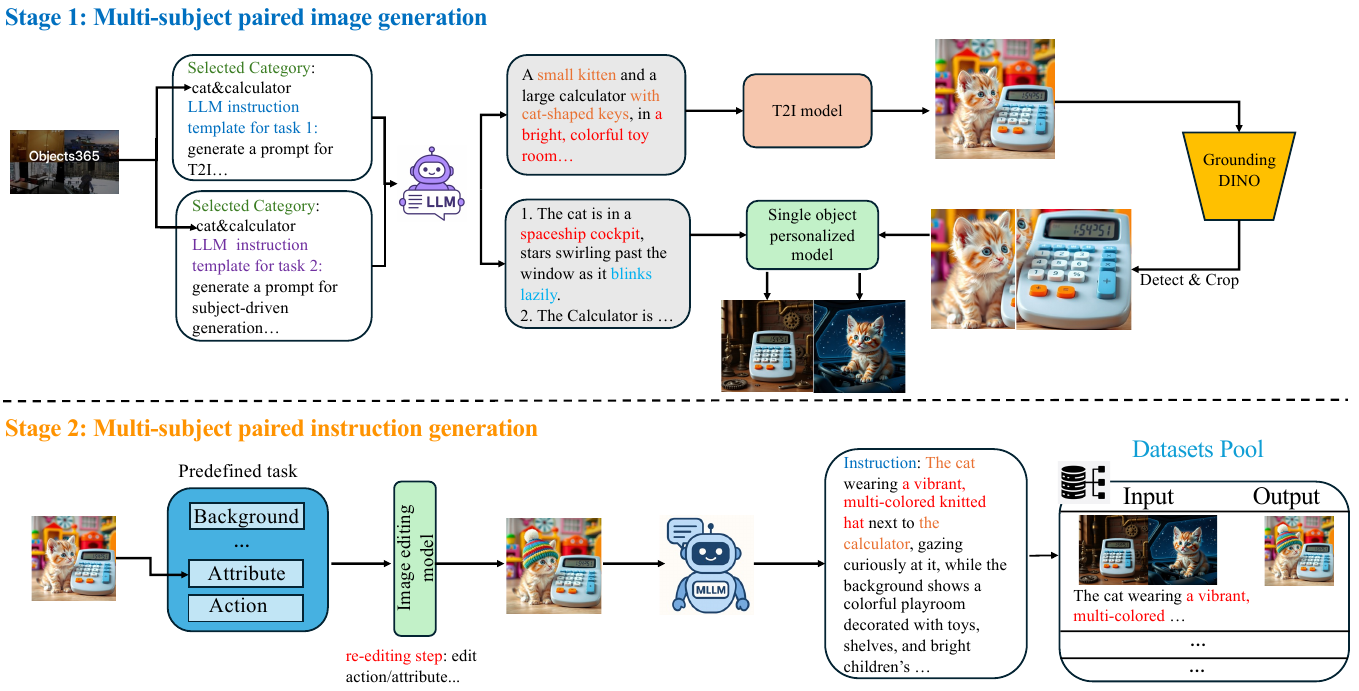}
\caption{Overview of the dataset construction pipeline. The process consists of two stages: (1) multi-subject paired image generation, and (2) multi-subject paired instruction generation.}
\label{fig:datasets-pipeline}
\vspace{-0.3cm}
\end{figure*}
\label{sec:Related Works}
\subsection{Multi-subject Driven Generation}
Personalized generation~\cite{custom-diffusion,ruiz2023dreambooth,shen2025long,shen2024imagpose,shen2025imagdressing, han2025show} refers to the task of synthesizing images of a target subject in novel scenes, given reference images containing that subject. Existing approaches to this task can be broadly categorized into two paradigms: test time fine-tuning and zero-shot methods. Early approaches~\cite{custom-diffusion,ruiz2023dreambooth,elite} such as DreamBooth~\citep{ruiz2023dreambooth} and Textual Inversion~\citep{gal2022image} adapt a model by fine-tuning selected parameters on 3–5 reference images of the subject, enabling the model to capture the desired concept. However, this paradigm requires training a separate model for each concept, incurring substantial computational and storage costs.

More recently, research has shifted toward zero-shot methods~\citep{ipadapter,photomaker,instantid,pulid,realcustom,realcustom++,ominicontrol,instantcharacter}, which aim to perform personalization for arbitrary subjects without the need for model fine-tuning. While such single-subject personalization methods demonstrate promising results, their performance remains limited when multiple subjects are involved. To address the challenges of multi-subject personalization, UNO~\cite{wu2025less} leverages T2I models to construct data and perform co-evolution. XVerse~\cite{chen2025xverse} achieves subject control by transforming reference images into shifts for text-stream modulation. Despite these advances, existing approaches~\cite{wu2025omnigen2,msdiffusion,wu2025less} still struggle to ensure consistency and are difficult to scale to larger sets of reference images.

\subsection{Reinforcement Learning for Generation}
Reinforcement learning (RL)~\citep{guo2025deepseek,schulman2017proximal,rlhf} has emerged as an effective paradigm for alignment and has demonstrated strong potential in visual generation~\citep{imagereward,liang2024rich,T2i-r1,got-r1,xue2025dancegrpo,liu2025flow,visionreward, directly-finetune,aligning-reward-backp,video-reward,dpok,fan2023optimizing,pref-grpo}.
Previous approaches~\cite{aligning-reward-backp,directly-finetune} optimized generative models for alignment by backpropagating through the reward signal.
More recently, policy-based reinforcement learning methods~\cite{xue2025dancegrpo,liu2025flow} have been introduced, which directly compute policy gradients to optimize the policy.
These approaches have demonstrated notable improvements in enhancing semantic alignment, aesthetic preference, and other aspects of pre-trained generative models.
For autoregressive T2I models, T2I-R1~\cite{T2i-r1} integrates semantic-level and token-level chain-of-thought reasoning and employs diverse rewards to enhance text alignment during generation. In the case of flow matching based models, approaches such as Flow-GRPO~\citep{liu2025flow} and DanceGRPO~\citep{xue2025dancegrpo} leverage different forms of reward to fine-tune T2I models via RL, thereby significantly improving their capabilities in semantic fidelity, text rendering, and human preference alignment. However, applying reinforcement learning to multi-subject personalized generation remains an open challenge.
\section{Datasets and Benchmark}
\label{sec:Datasets and Benchmark}

The core of our dataset construction pipeline lies in leveraging the powerful capabilities of large language models, Text-to-Image generation models, and single-subject personalization models in a synergistic manner, enabling the creation of datasets that can scale to an arbitrary number of personalized subjects. Specifically, as shown in Figure~\ref{fig:datasets-pipeline}, our dataset construction pipeline consists of two stages: multi-subject paired image generation and multi-subject paired instruction generation.
\subsection{Scalable Multi-subject Driven Datasets }
\paragraph{Stage 1: multi-subject paired image generation.}

Existing approaches for synthesizing personalized generation data often rely solely on T2I models combined with strict subject-consistency filtering strategies, such as UNO~\citep{wu2025less} and Subjects200K~\citep{ominicontrol}, where a T2I model generates a diptych containing the same subject depicted across two different scenes. However, due to the inherent instability of T2I models in producing such data, these methods often yield datasets of limited consistency and quality.

In contrast, recent advances in single-subject personalization models~\cite{labs2025flux,qwen-image} have demonstrated strong capabilities: given a single subject, these models can reliably generate its appearance across novel scenes while maintaining high consistency. Motivated by the capability of these models, we propose a highly scalable multi-subject paired image generation pipeline that harnesses the strengths of these powerful models.
Specifically, as shown in Figure~\ref{fig:datasets-pipeline}, for constructing data with $n$ subjects, we first sample an n-element category set $\mathcal{C} = \{c_1, c_2, \dots, c_n\}$ from the full category pool of Objects365~\citep{objects365}. We then prompt a large language model (LLM)~\cite{qwen3} to generate a text-to-image instruction $T^{t2i}$, where each category’s appearance is explicitly specified to increase data diversity. In addition, we prompt the LLM~\cite{qwen3} to produce subject-driven instructions $T^{sub}$ for each category, which will be used for generating single subject personalization images. Using a state-of-the-art T2I models~\cite{flux,qwen-image}, we employ $T^{t2i}$ to synthesize multi-subject images $I_{out}$. Then we leverage GroundingDINO~\citep{groundingDino}, a grounding-based object detection model, to detect and crop instances by category in $I_{out}$, resulting in single-subject images $I_{crop}$. Finally, given $I_{crop}$ and $T^{sub}$, we generate new reference images $I_{ref}$ for each subject using the single-image personalization models~\cite{labs2025flux}, which serve as the input references of our multi-subject generation dataset. This stage enables the construction of diverse and high-quality image pairs.

\vspace{-0.5cm}
\paragraph{Stage 2: multi-subject paired instruction generation.}
In Stage 1, to guide the T2I model~\cite{flux,qwen-image} toward producing higher-quality and more distinctive images, the prompts include explicit descriptions of the subjects’ appearances. However, directly reusing these descriptions for multi-subject personalization may cause the model to exploit textual leakage: focusing only on the appearance information revealed in the text rather than attending to the reference images themselves. Furthermore, to enhance the diversity of our dataset, we hope to ensure that these multi-subject paired instructions cover a broader range of tasks, such as instructions focused on describing spatial relationships or those capable of detailing the attributes of each subject.

To address these challenges, we first predefine seven types of multi-subject personalized tasks, including attribute personalization, background personalization, action personalization, positional personalization, complex-scene personalization, three-subject personalization, and four-subject personalization. More details about these tasks are provided in the Supplementary Material. For each task, we apply recaptioning strategies, as shown in Figure~\ref{fig:datasets-pipeline}. For example, in the background personalization task, we prompt MLLM~\cite{qwen25vl} to focus on describing the background of $I_{out}$ while explicitly using pronouns to refer to specific subjects, thus avoiding the leakage of subject appearance information within the instructions. Specifically, for tasks involving attributes or actions, the subject images generated in Stage 1 may not exhibit sufficiently prominent attributes. Therefore, we introduce an additional `re-editing' step to modify the original $I_{out}$. Concretely, we first use an LLM~\cite{qwen3} to generate task-specific editing instructions. These instructions, together with $I_{out}$, are then fed into an editing model to produce $I_{out}^*$. Similarly, we use $I_{out}^*$ to prompt MLLM~\cite{qwen25vl} to generate the corresponding instructions. This stage yields a large set of high-quality, diverse, and personalized instruction data, which substantially enriches the dataset and supports more effective downstream training.
\subsection{PSRBench}
Existing benchmarks for evaluating multi-subject personalization exhibit significant limitations in both evaluation tasks and metrics. For example, DreamBench~\cite{ruiz2023dreambooth} only includes combinations of two subjects with overly simplistic scene descriptions, and evaluates the semantic consistency of images using CLIP~\cite{clip}, which can not accurately capture the true semantic alignment. Moreover, current evaluation protocols are often coarse-grained. For instance, UNO~\cite{wu2025less} computes the DINO score~\cite{dinov2} between each subject in a generated two-subject image and its corresponding reference image, but such a method cannot provide a fine-grained assessment of subject consistency. OmniContext~\cite{wu2025omnigen2} leverages GPT-4.1 for evaluation, yet it still fails to reliably measure consistency. XVerse~\cite{chen2025xverse} proposes a segmentation-based approach, where each subject in the generated image is segmented and then compared to the corresponding input reference using DINO scores~\cite{dinov2}. However, this method assumes that reference images contain isolated subjects on plain white backgrounds—an idealized setting that does not generalize well to real-world scenarios, where reference images typically include subjects within simple or even complex backgrounds.
\begin{figure*}[t]
\centering
\includegraphics[width=0.95\linewidth]{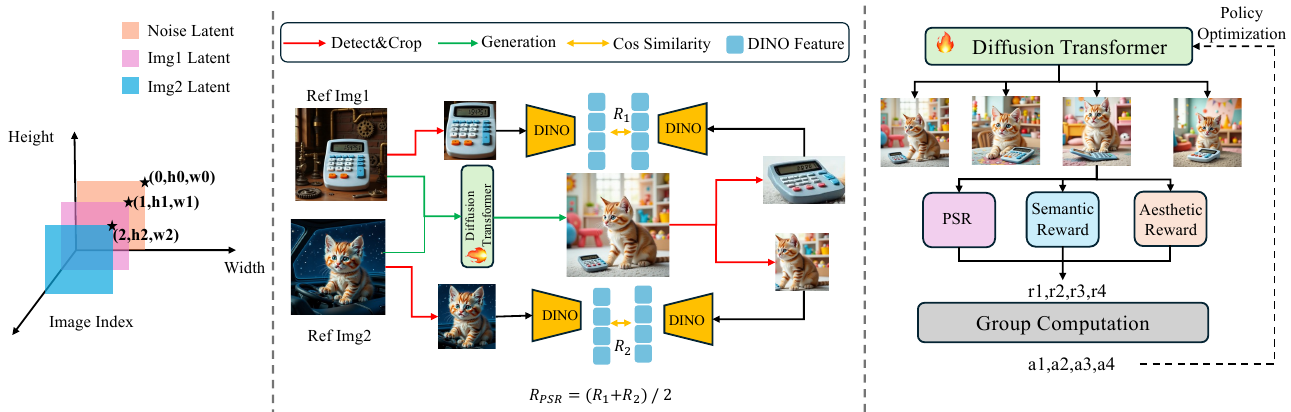}
\vspace{-0.2cm}  
\caption{Left: Scalable frame-wise positional encoding. Middle: Pairwise subject-consistency rewards. Right: GRPO training pipeline combining PSR with multiple rewards.}
\label{fig:methods_b}
\end{figure*}

To this end, we propose PSRBench, a comprehensive and multi-dimensional benchmark for multi-subject personalization. Specifically, our benchmark consists of seven subsets: Attribute, Background, Action, Position, Complex, Three, and Four, each representing a distinct sub-task. Each subset is evaluated along three complementary dimensions: subject consistency, semantic alignment, and aesthetic preference. For subject consistency, we adopt a grounding-based~\cite{groundingDino} approach: both input and output images are first processed with an object grounding model~\cite{groundingDino} to detect and crop the subjects, after which DINO scores~\cite{dinov2} are computed on corresponding subject pairs. This design enables precise and fine-grained evaluation of subject consistency. For semantic alignment, we first employ an MLLM~\citep{qwen25vl} to evaluate image-text alignment according to the specific task requirements. To more precisely assess positional alignment, a grounding-based evaluation is additionally performed on the Position subset, wherein the centers of the subjects are localized and their relative spatial relations are compared against the prompt. For aesthetic preference, we rely on HPSv3~\citep{ma2025hpsv3} for evaluation.

\section{Methods}
\label{sec:Methods}
\subsection{Preliminary}
\paragraph{Flow Matching}

Flow Matching model is trained by minimizing the objective:
\begin{equation}
\min_v \int_0^1 \mathbb{E}\left[\left| (z_1 - z_0) - v(z_t,t) \right|^2 \right] dt
\label{eq:4}
\end{equation}

During the sampling process, the model starts from random noise, and the ODE is solved using a simple Euler solver:
\begin{equation}
z_{t_{i-1}} = z_{t_i} + v(z_t,t)\Delta t
\label{eq:5}
\end{equation}








\paragraph{GRPO for flow matching}

Flow-GRPO~\citep{liu2025flow} transforms the ODE-based sampling into an SDE form to incorporate the stochasticity required by GRPO. Through Euler-Maruyama discretization, the resulting policy update is as follows:
\begin{multline}
    x_{t+\Delta t}=x_t+\Big(v_\theta\left(x_t, t\right)
+\frac{\sigma_t^2}{2 t}\big(x_t+(1-t) v_\theta\left(x_t, t\right)\big)\Big)\Delta t \\
+\sigma_t \sqrt{\Delta t} \epsilon, 
\quad \epsilon \sim \mathcal{N}(0, I)
\end{multline}
Flow-GRPO~\citep{liu2025flow} set $\sigma_t=a \sqrt{\frac{t}{1-t}}$, $a$ is a scalar hyper-parameter that controls the noise level.
\subsection{Scalable Frame-wise Positional Encoding}
In the SFT stage, our goal is to endow a single-image personalization model with the knowledge required to handle multi-image, multi-subject scenarios. To achieve this, we leverage a frame-wise positional offset together with a multi-image joint training strategy.
Similar to prior work~\citep{wu2025less,labs2025flux,ominicontrol}, we also employ a VAE encoder to encode input images, and then concatenate the resulting latents with the noise latent. In the case of multi-image inputs, this token-level concatenation remains a natural choice; however, it requires a specific positional indexing scheme to distinguish tokens from different input images. Some previous approaches~\cite{wu2025less} introduced offsets along the $h$ and $w$ dimensions. For example, UNO~\cite{wu2025less} suggested that the spatial index of the $i^{th}$ image should start from the terminal position of the $(i-1)^{th}$ image. While such indexing allows better utilization of pretrained model capacity, it suffers from two key limitations. First, by enforcing offsets along the $h$ and $w$ dimensions, the model implicitly inherits a strong prior that the second image is naturally positioned to the right or below the first image, which complicates fine-grained control through textual prompts. Second, when scaling to more images, e.g., three or four inputs, such large spatial offsets in $h$ and $w$ become increasingly misaligned with the pretrained distribution, ultimately degrading the model’s generalization ability.

Therefore, as shown in Figure~\ref{fig:methods_b}, we extend the positional encoding scheme proposed in FLUX.1 Kontext~\cite{labs2025flux} by employing only a virtual temporal offset to indicate the index of each input image. Concretely, for the latent tokens of the $i^{th}$ input image, the positional offset is defined as
\begin{equation}
    PO_i=(i,h,w),
\end{equation}
where $h,w$ denote the latent’s spatial dimensions.
During training, we adopt a multi-image joint training strategy that incorporates datasets with varying numbers of input images. We argue that jointly training across different reference counts allows the model to exploit their complementary benefits, leading to improved generalization and robustness in multi-subject personalization.
\begin{table*}[t]
\captionsetup{skip=2pt}
    \centering
    \caption{Quantitative comparison of subject consistency on PSRBench. ``Ours-SFT" denotes the model trained with only the first-stage supervised fine-tuning. ``Ours (PSR)" represents the models further trained using the proposed reinforcement strategy. }
    \label{tab:sc_results}
    \resizebox{0.95\textwidth}{!}{
    \begin{tabular}{lcccccccc}
        \toprule
        Model & Attribute & Background & Action & Position & Complex & Three & Four & \textbf{Overall} \\
        \cmidrule(lr){1-1}
        \cmidrule(lr){2-8}
        \cmidrule(lr){9-9}
FLUX.1 Kontext~\citep{labs2025flux} & 0.510 & 0.534 & 0.535 & 0.560 & 0.543 & 0.425 & 0.373 & 0.497 \\        
UNO~\citep{wu2025less} & 0.532 & 0.568 & 0.573 & 0.584 & 0.540 & 0.460 & 0.402 & 0.523 \\
OmniGen2~\citep{wu2025omnigen2} & 0.602 & 0.559 & 0.628 & 0.678 & 0.610 & 0.527 & 0.508 & 0.587 \\
XVerse~\citep{chen2025xverse} & 0.599 & 0.624 & 0.694 & 0.595 & 0.619 & 0.541 & 0.435 & 0.587 \\
Qwen-Image-Edit-2509~\citep{qwen-image} & 0.593 & 0.539 & 0.634 & 0.653 & 0.546 & 0.552 & 0.361 & 0.554 \\
\midrule
Ours-SFT & 0.551 & 0.573 & 0.631 & 0.582 & 0.587 & 0.519 & 0.472 & 0.559 \\
\textbf{Ours (PSR)} & \textbf{0.626} & \textbf{0.721} & \textbf{0.741} & \textbf{0.727} & \textbf{0.713} & \textbf{0.615} & \textbf{0.571} & \textbf{0.673} \\
        \bottomrule
    \end{tabular}
    }
    \vspace{-0.2cm}
\end{table*}
\subsection{Pairwise Subject-Consistency Rewards}
Despite the benefits of the first-stage SFT, two key challenges remain in multi-subject personalization. On the one hand, while SFT enables the model to process multiple reference images, its optimization operates only at the global image level and therefore lacks fine-grained supervision for individual subjects. Personalized generation inherently requires subject-level alignment, which is difficult to achieve through standard SFT loss. RL~\cite{xue2025dancegrpo,liu2025flow}, in contrast, provides a mechanism to directly optimize localized objectives via subject-aware rewards. Moreover, in multi-subject personalization scenarios~\cite{wu2025less,msdiffusion}, the reference images often contain background elements, making it essential for the model to robustly extract and preserve subject identity under realistic inputs. 

To this end, we introduce a novel Pairwise Subject-Consistency Reward (PSR), as illustrated in Figure~\ref{fig:methods_b}, and employ an online RL stage to post-train the model initialized from the first-stage. The key idea behind PSR is subject decoupling: we disentangle each subject from the global image, and then encourage pairwise similarity between the disentangled subjects and their references to guide training.
Formally, let a pretrained model 
$\theta$ take multiple input images 
, each containing multiple subjects, and produce an output $I_{out}$. We apply subject decoupling to the output to obtain subject-specific crops:
$I_{dec}^i=g(I_{out},c_i)$, where $g$ denotes an open-vocabulary object detector and $c_i$ represents the category of the $i^{th}$ subject. The decoupled image 
$I_{dec}^i$ contains only the subject-specific region. Since input references in real-world scenarios also contain background, we perform the same decoupling operation on input references:
$I_{gt}^i=g(I_{ref}^i,c_i)$
We then define the subject consistency reward as the average similarity between each pair of corresponding subject crops:
\begin{equation}
R_{PSR}=\frac{1}{N} \sum^{N}_{i=1} f\left(I_{dec}^i, I_{gt}^i\right)
\end{equation}
where $N$ denotes the number of input images, and $f$ measures visual similarity using the DINO features~\cite{dinov2}.

To further mitigate the risk of copy-paste artifacts and prevent reward hacking, we further incorporate aesthetic preference and semantic alignment rewards. The overall reward is defined as follows:

\begin{equation}
    R=w_1*R_{PSR}+w_2*R_s+w_3*R_h
\end{equation}
where $w_1,w_2,w_3$ denote the weights of different reward terms, and $R_s$ is the semantic reward computed by Qwen2.5-VL~\citep{qwen25vl}:
\begin{equation}
    R_s=MLLM(instruction,I_{out}),
\end{equation}
$R_h$ is aesthetic preference reward from a human preference scoring model HPSv3~\citep{ma2025hpsv3}:
\begin{equation}
    R_h=HPS(instruction,I_{out}),
\end{equation}


\begin{table*}[t]
\captionsetup{skip=2pt}
    \centering
    \caption{Quantitative comparison of aesthetic preference on PSRBench.}
    \label{tab:hps_results}
    \resizebox{0.95\textwidth}{!}{
    \begin{tabular}{lcccccccc}
        \toprule
        Model & Attribute & Background & Action & Position & Complex & Three & Four & \textbf{Overall} \\
        \cmidrule(lr){1-1}
        \cmidrule(lr){2-8}
        \cmidrule(lr){9-9}
FLUX.1 Kontext~\citep{labs2025flux} & 0.811	& 0.968&	0.776&	0.828&	0.967&	0.886&	0.857 & 0.870 \\
UNO~\citep{wu2025less} &0.876	&1.060	&\textbf{0.960}	&1.060	&0.979	&1.100	&1.030 & 1.009 \\
OmniGen2~\citep{wu2025omnigen2} &0.908	&1.100	&0.911	&1.070	&0.983	&1.070	&1.100& 1.020 \\
XVerse~\citep{chen2025xverse} &0.882	&0.936	&0.754	&0.970	&0.864	&0.938	&0.909 & 0.893 \\
Qwen-Image-Edit-2509~\citep{qwen-image}&0.883	&0.890	&0.877	&0.961	&0.898	&0.981	&0.533& 0.860 \\
\midrule
Ours-SFT &0.644	&0.819	&0.634	&0.779	&0.771	&0.952	&0.961& 0.794 \\
\textbf{Ours (PSR)}  &\textbf{1.040}	&\textbf{1.150}	&0.881	&\textbf{1.200}	&\textbf{1.050}	&\textbf{1.260}	&\textbf{1.290} & \textbf{1.124}\\
        \bottomrule
    \end{tabular}
    }
    \vspace{-0.2cm}
\end{table*}
\begin{table*}[h]
\captionsetup{skip=2pt}
    \centering
    \caption{Quantitative comparison of semantic alignment on PSRBench.}
    \label{tab:semantic_results}
    \resizebox{0.95\textwidth}{!}{
    \begin{tabular}{lcccccccc}
        \toprule
        Model & Attribute & Background & Action & Position & Complex & Three & Four & \textbf{Overall} \\
        \cmidrule(lr){1-1}
        \cmidrule(lr){2-8}
        \cmidrule(lr){9-9}
FLUX.1 Kontext~\citep{labs2025flux} &0.710	&0.810	&0.600	&0.058	&0.644	&0.656	&0.605& 0.583 \\
UNO~\citep{wu2025less} &0.701	&0.864	&0.662	&0.250	&0.740	&0.746	&0.704& 0.667 \\
OmniGen2~\citep{wu2025omnigen2} &0.833	&0.908	&\textbf{0.895}	&0.389	&\textbf{0.768}	&0.784	&0.727& 0.758 \\
XVerse~\citep{chen2025xverse} &0.758	&0.839	&0.648	&0.259	&0.689	&0.779	&0.709& 0.669 \\
Qwen-Image-Edit-2509~\citep{qwen-image}&\textbf{0.918}	&\textbf{0.945}	&0.826	&0.439	&0.713	&\textbf{0.894}	&0.592& 0.761 \\
\midrule
Ours-SFT &0.822	&0.916	&0.709	&0.310	&0.640	&0.828	&0.762& 0.712 \\
\textbf{Ours (PSR)}  &0.908	&0.926	&0.739	&\textbf{0.468}	&0.692	&0.884	&\textbf{0.866}& \textbf{0.783} \\
        \bottomrule
    \end{tabular}
    }
    \vspace{-0.3cm}
\end{table*}

\section{Experiments}
\label{sec:experiments}
\begin{figure*}[t]
\centering
\includegraphics[width=0.95\linewidth]{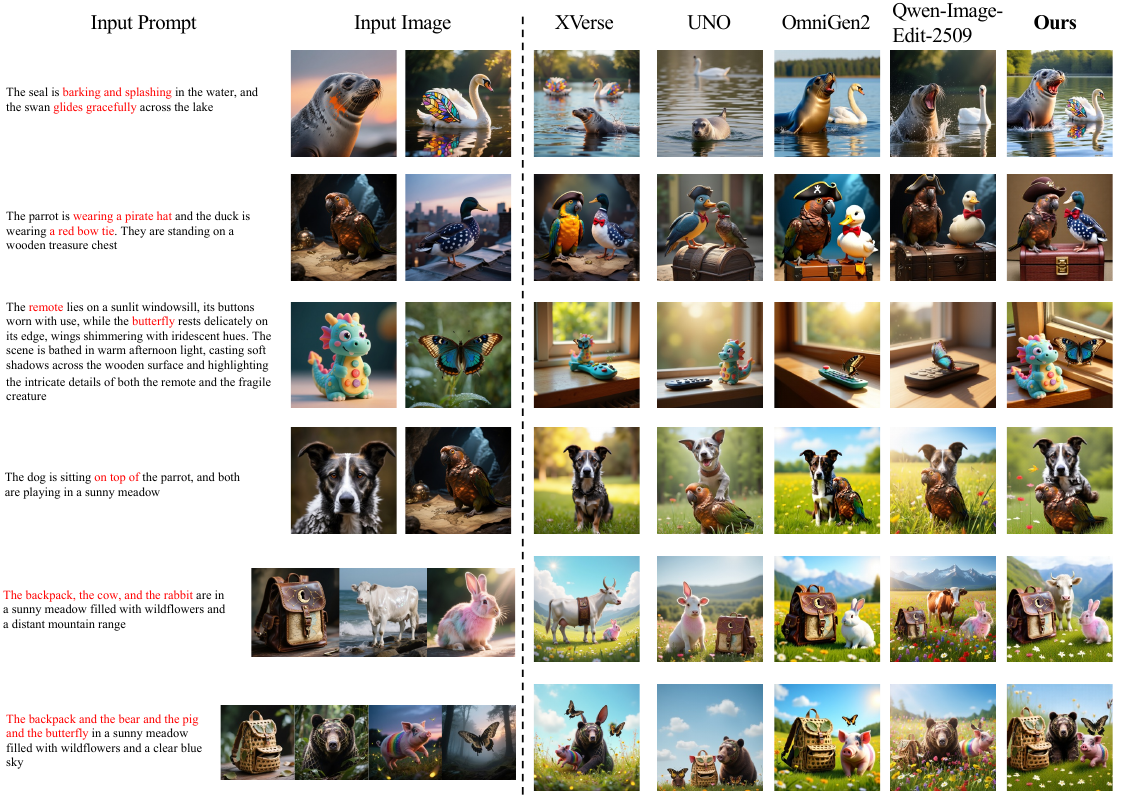}
\caption{Qualitative analysis results of PSR with recent state-of-the-art models. }
\label{fig:qual}
\vspace{-10pt}
\end{figure*}
\subsection{Implementation Details}
We build our multi-subject personalization model on top of FLUX.1 Kontext~\cite{labs2025flux} and train the model using LoRA~\cite{lora}. In the first stage, we set the learning rate to 1e-4. Our dataset provides 2–4 reference images. During training, we sample 2, 3, and 4 references with probabilities of 0.9, 0.05, and 0.05, respectively, enabling joint training across different reference counts. In the second stage, we use a smaller learning rate of 
1e-5. 
Specifically, we adopt a joint reinforcement training strategy with multiple rewards. The reward for subject consistency is provided by our proposed PSR. For textual faithfulness, we leverage Qwen2.5-VL-32B-Instruct~\cite{qwen25vl} to evaluate image-text alignment, while aesthetic preference is modeled using HPSv3~\cite{ma2025hpsv3}.
Since this stage requires evaluating human preferences, sampling and training are both conducted over the original 28 diffusion timesteps. All square images are resized to a resolution of 512×512. For non-square images, the shortest side is proportionally resized to 512 pixels while maintaining the original aspect ratio.


We use the model trained with only the first-stage, referred to as Ours-SFT, as the baseline. We also compare our proposed PSR with state-of-the-art multi-subject personalization models, including UNO~\citep{wu2025less}, OmniGen2~\citep{wu2025omnigen2}, XVerse~\cite{chen2025xverse}, and FLUX.1 Kontext~\cite{labs2025flux}. Since FLUX.1 Kontext~\cite{labs2025flux} only supports single-image inputs, for fair evaluation we concatenate multiple input images along the width dimension before feeding them into the model. We also compared with the latest Qwen-Image-Edit-2509~\cite{qwen-image}, which supports multi-image input.


\subsection{Comparisons with Recent Methods}
\label{sec:quan}
PSRBench employs three distinct metrics to evaluate subject consistency, semantic alignment, and human preference. The detailed calculation for these metrics can be found in the Supplemental Material. Table~\ref{tab:sc_results} presents a detailed comparison of subject consistency across seven subsets of PSRBench. Our method achieves the best performance not only in terms of the overall score (0.673), but also consistently outperforms all baselines across every individual subset. In particular, the improvements are especially pronounced in more challenging scenarios involving complex prompts and interactions among more subjects. For example, on the Three and Four subsets, our model reaches 0.615 and 0.571, respectively, substantially surpassing the previous best results of 0.552 and 0.508. These gains indicate that our model is more robust in handling consistency entanglement when multiple images are provided as input, which is a common failure mode of existing methods.


\vspace{-0.1cm}
Table~\ref{tab:hps_results} and Table~\ref{tab:semantic_results} report the aesthetic preference score and semantic alignment score, respectively. PSR also achieves the overall best performance on these two metrics, indicating that our method not only preserves consistency in multi-subject scenarios but also exhibits strong text controllability, enabling the generation of high-quality images. Notably, compared with baseline model FLUX.1 Kontext~\cite{labs2025flux}, “Ours-SFT” exhibits a drop in aesthetic scores. We attribute this decline to the mismatch in training resolution: the original FLUX.1 Kontext~\cite{labs2025flux} is trained at 1024 resolution, whereas our SFT stage is primarily conducted at 512 resolution. This downscaling disrupts the generative priors of the original model and consequently degrades the visual quality, leading to lower aesthetic preferences. However, our PSR-based reinforcement learning effectively mitigates this issue and restores the overall perceptual quality. Furthermore, benefiting from our multi-reward joint optimization, our method achieves notably superior performance on the Position and Four subsets in the semantic alignment evaluation. Despite their difficulty, our approach shows clear superiority on these subsets.


\subsection{Qualitative Analysis}
\label{sec:qual}
Figure~\ref{fig:qual} presents qualitative analyses on PSRBench. In comparison with other baselines, our approach demonstrates superior performance in both subject preservation and text adherence. In the case of two-image inputs, our method is able to faithfully retain the original appearance of the subjects while simultaneously complying with semantic instructions, thereby validating the effectiveness of both our data and methodology. For example, our approach successfully preserves the distinctive visual characteristics of the ``remote" and ``butterfly", while also adhering to the textual description that specifies ``sunlit windowsill". In contrast, other methods often fail to maintain high consistency across multiple subjects, with XVerse~\cite{chen2025xverse} even suffering from complete subject omission. Furthermore, our method demonstrates powerful text alignment capabilities while maintaining strong subject consistency: when the prompt contains positional, action, or attribute-binding constraints, our method can still generate coherent images. While other models such as OmniGen2~\cite{wu2025omnigen2} and Qwen-Image-Edit-2509~\cite{qwen-image} also exhibit robust overall performance in text adherence, they notably suffer from diminished subject consistency.


\subsection{Ablation Study}
We further conduct ablation studies on our proposed scalable frame-wise positional encoding to validate its effectiveness. Specifically, we hypothesize that existing encoding schemes introduce inherent spatial positional priors, which in turn restrict the model's capacity to understand and represent semantics that deviate from this intrinsic prior. To verify this, we randomly sample 50K training samples and train each variant with a different positional encoding for one epoch. We then evaluate their semantic alignment scores on 2-subject, 3-subject, and 4-subject settings, as well as the Position subset.
As shown in Table~\ref{tab:ablation1}, our method consistently achieves the best semantic alignment across multi-subject personalization tasks. Notably, on the Position subset, our approach demonstrates a significant lead, outperforming the second-best method by $0.39$. We attribute this to the fact that alternative encoding methods impose fixed spatial layouts, making it harder for the model to modify subject positions according to the textual prompt. In contrast, our frame-wise positional encoding operates solely along the frame dimension, effectively avoiding such pre-imposed spatial biases.
\begin{table}[t]
\captionsetup{skip=2pt}
\centering
\setlength{\tabcolsep}{4pt} 
\caption{Comparison of different offset strategies for the positional index. The notation ``w/ h-w" indicates that the positional indexing is offset in both the height and width dimensions. Conversely, ``w/ w" denotes an offset applied solely along the width dimension, and ``w/ h" signifies an offset applied solely along the height dimension. We compare the semantic scores across different subsets. Our positional encoding strategy (``w/ ours") achieves overall superior performance, with particularly notable improvements on the Position subset.}
\label{tab:ablation1}
\begin{tabular}{lcccc}
\toprule
\textbf{Method} & \textbf{2 subjects} & \textbf{3 subjects} & \textbf{4 subjects}& \textbf{Position} \\
\midrule
w/ h-w         & \textbf{0.929} & 0.831 & 0.808 & 0.469 \\
w/ w    & 0.915 & 0.824 & 0.777 & 0.459 \\
w/ h   & 0.925 & 0.840 & 0.805 & 0.437\\
\midrule
\textbf{w/ ours}      & 0.922 & \textbf{0.870} & \textbf{0.821} &  \textbf{0.508} \\
\bottomrule
\end{tabular}
\vspace{-0.7cm}
\end{table}


\vspace{-0.2cm}
\section{Conclusion}
\label{sec:conclusion}
In this work, we addressed the challenges of multi-subject personalized image generation, where existing models struggle to maintain subject consistency and semantic alignment. To overcome these limitations, we proposed a scalable multi-subject data generation pipeline and introduced Pairwise Subject-Consistency Rewards within a multi-reward RL framework. Furthermore, we designed PSRBench, a comprehensive benchmark that evaluates subject consistency, aesthetic preference, and semantic alignment across diverse scenarios. Experiments confirm our SOTA performance in consistency and controllability, yiel ding high-fidelity, semantically faithful results. This work provides a solid foundation for future research on controllable and scalable multi-subject personalized generation.



\clearpage
\setcounter{page}{1}
\maketitlesupplementary
\begin{table}[t]
\captionsetup{skip=2pt}
\centering
\setlength{\tabcolsep}{4pt} 
\caption{Quantitative Comparison on DreamBench}
\label{tab:dreambench}
\begin{tabular}{lccc}
\toprule
\textbf{Method} & \textbf{CLIP-T} & \textbf{CLIP-I} & \textbf{DINO} \\
\midrule
UNO~\cite{wu2025less}         & 0.319 & 0.713 & 0.492 \\
OmniGen2~\cite{wu2025omnigen2}   & 0.329 & \textbf{0.720} & 0.506 \\
XVerse~\cite{chen2025xverse}   & 0.327 & 0.715 & 0.479\\
Qwen-Image-Edit-2509~\cite{qwen-image}   & \textbf{0.337} & 0.716 & 0.494\\
\midrule
\textbf{Ours (PSR)}      & 0.335 & 0.717 & \textbf{0.529} \\
\bottomrule
\end{tabular}
\vspace{-0.2cm}
\label{tb:dreambench}
\end{table}

\section{More Implementation Details}
We adopt LoRA~\cite{lora} in both training stages, using a rank of 512 in the first stage and 64 in the second stage. During the GRPO training phase, we set the group size to 6, which yields stable convergence in practice.
Since the aesthetic preference reward cannot be naturally normalized to the $[0,1]$ range and is often larger than the other rewards in our experiments, we adjust the weighting scheme to stabilize multi-reward training. Specifically, we assign weights $w_1=0.4, w_2=0.4$ and $w_3=0.2$ to the respective reward components.


\section{Experimental Results on DreamBench}
To further validate the effectiveness of our approach, we also conduct evaluations on DreamBench~\cite{ruiz2023dreambooth,wu2025less}. Specifically, we randomly sample 100 test cases from the DreamBench~\cite{ruiz2023dreambooth,wu2025less} multi-ip subset and compare our method against recent state-of-the-art approaches, including UNO~\cite{wu2025less}, OmniGen2~\cite{wu2025omnigen2}, XVerse~\cite{chen2025xverse}, and Qwen-Image-Edit-2509~\cite{qwen-image}. The evaluation follows the official testing protocol provided by UNO~\cite{wu2025less}.
As shown in Table~\ref{tab:dreambench}, our method achieves state-of-the-art performance on the DINO metric, surpassing the second-best method on DINO by a margin of 0.23. In addition, our model attains competitive results with Qwen-Image-Edit-2509~\cite{qwen-image} on the CLIP-T metric. These findings collectively demonstrate the superiority of our approach. Although our method outperforms existing approaches on these metrics, it is important to acknowledge that there are still limitations in this evaluation method. For example, CLIP's semantic evaluation is not always accurate, and the global DINO score for assessing subject consistency is susceptible to interference from background information. We will discuss these issues in further sections.

\begin{figure*}[h]
\includegraphics[width=1.0\linewidth]{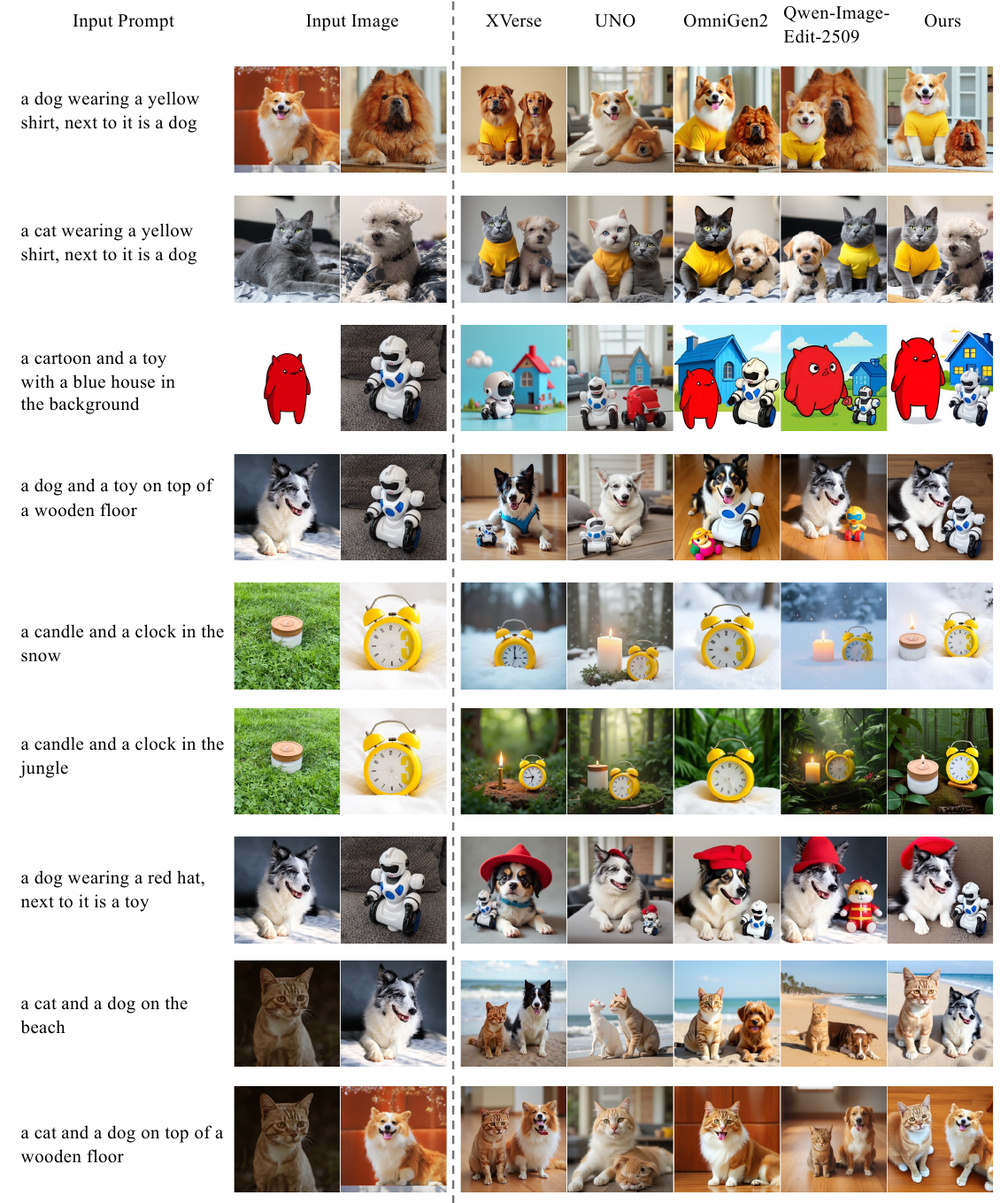}
\caption{Qualitative Analysis on DreamBench}
\label{fig:dreambench}
\end{figure*}

In addition, Figure~\ref{fig:dreambench} presents the qualitative results of our model on the DreamBench~\cite{ruiz2023dreambooth}. Consistent with the observations on PSRBench, our method demonstrates superior subject consistency compared to other approaches. For instance, in the fourth row, existing methods fail to preserve the appearance of the robot toy in the generated images. In contrast, our method performs exceptionally well in maintaining both the dog's and the robot's appearances. Furthermore, as shown in the fifth row, when the input image features a candle that is not similar to a conventional candle, existing methods either disregard this 'non-standard' candle or generate one that is completely dissimilar to the original. Our method, on the other hand, handles such challenging cases effectively.

\begin{table*}[h]
    \centering
    \tiny
    \caption{Quantitive Comparison with More Baseline Models}
    \setlength{\arrayrulewidth}{0.3pt} 
    \vspace{-11pt} 
    \resizebox{1\linewidth}{!}{ 
        \begin{tabular}{l c c c c c c c c}  
            \hline 
            Method & Attribute & Background & Action& Position&Comlex&Three&Four&Overal  \\
            \hline
            MS-Diffusion (SC)~\cite{msdiffusion}    & 0.352   & 0.366  & 0.401 & 0.329   & 0.322  & 0.303 & 0.300   & 0.339 \\
            MOSAIC(SC)~\cite{she2025mosaic}  & 0.569   & 0.505  & 0.580 & 0.555   & 0.513  & 0.419 & 0.366   & 0.501 \\
            \textbf{Ours (SC)}   & \textbf{0.626}   & \textbf{0.721}  & \textbf{0.741} & \textbf{0.727}   & \textbf{0.713}  & \textbf{0.615} & \textbf{0.571}   & \textbf{0.673} \\
            \hline
            MS-Diffusion (HPS)~\cite{msdiffusion}   & 0.659   & 0.930  & 0.704 & 0.865   & 0.774  & 0.977 & 0.974   & 0.840 \\
            MOSAIC (HPS)~\cite{she2025mosaic}   & 0.922   & 1.106  & 0.933 & 1.116   & 0.964  & 1.107 & 0.997   & 1.021 \\
            \textbf{Ours (HPS)}   & \textbf{1.040}   & \textbf{1.150}  & \textbf{0.881} & \textbf{1.200}   & \textbf{1.050}  & \textbf{1.260} & \textbf{1.290}   & \textbf{1.124} \\
            \hline
            MS-Diffusion (SA)~\cite{msdiffusion}   & 0.587   & 0.849  & 0.479 & 0.254   & 0.706  & 0.692 & 0.684   & 0.607 \\
            MOSAIC (SA)~\cite{she2025mosaic}    & 0.776   & 0.921  & 0.693 & 0.274   & \textbf{0.818}  & 0.827 & 0.786   & 0.728 \\
            \textbf{Ours (SA)}   & \textbf{0.908}   & \textbf{0.926}  & \textbf{0.739} & \textbf{0.468}   & 0.692  & \textbf{0.884} & \textbf{0.866}   & \textbf{0.783} \\
            \hline
        \end{tabular}
    }
    \label{tb:more baseline}
\end{table*}

\begin{table}[h]
    \centering
    \caption{User Study}
    \setlength{\arrayrulewidth}{0.3pt} 
    \resizebox{1.0\linewidth}{!}{ 
        \begin{tabular}{l c c c}  
            \hline 
            Method & SC & SA & HPS  \\
            \hline
            Omnigen2~\cite{wu2025omnigen2}  &0.50  &0.58  &0.74   \\
            XVerse~\cite{chen2025xverse}  &0.64  &0.66  &0.60   \\
            Qwen-Image-Edit-2509~\cite{qwen-image}    &0.74    &0.76   &0.62   \\
            \textbf{Ours (PSR)}    &\textbf{0.92}    &\textbf{0.80}   &\textbf{0.82}   \\
            \hline
        \end{tabular}
    }
    \label{tb:user}
\end{table}

\begin{figure*}[h]
\includegraphics[width=1.0\linewidth]{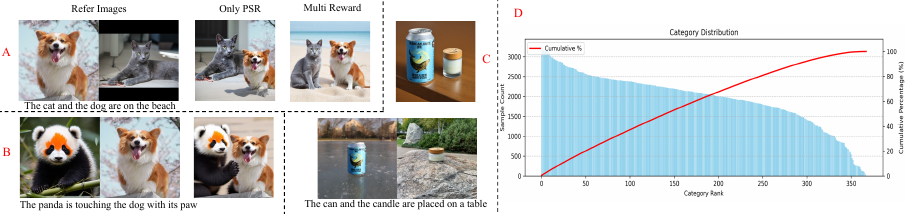}
\caption{(A) Hacking hacking. (B) Interaction results. (C) Failure case. (D) Data statistics.}
\label{fig:fig-all}
\end{figure*}

\section{Quantitive Comparison with More Baseline}
To further validate the effectiveness of our approach, we benchmark it against MS-Diffusion~\cite{msdiffusion} and the state-of-the-art MOSAIC~\cite{she2025mosaic}. As shown in Table~\ref{tb:more baseline}, our method consistently outperforms these baselines across all metrics. Specifically, SC, HPS, and SA denote Subject Consistency, Aesthetic Score, and Semantic Alignment, respectively.

\section{User Study}
Tab.~\ref{tb:user} reports the user study results. Specifically, we randomly sample 100 test cases and ask five participants to rank images generated by these four methods for each case, where the ranked images are assigned scores of 1, 0.8, 0.6, and 0.4 from highest to lowest. The results indicate that our method achieves the best performance. 

\section{Additional Visualization Results and Analysis}
As shown in Fig.~\ref{fig:fig-all}-A, combining multiple rewards effectively mitigates reward hacking, whereas PSR alone tends to induce copy-and-paste behavior.
Fig.~\ref{fig:fig-all}-B presents cases where our model successfully captures inter-subject interactions, demonstrating that the results go beyond simple subject stacking and exhibit overall coherence. Fig.~\ref{fig:fig-all}-C exemplifies a notable failure case of our method, specifically illustrating that identity preservation struggles to be maintained for small subjects.


\section{More Infomation about Datasets and PSRBench}
\begin{figure*}[h]
\includegraphics[width=1.0\linewidth]{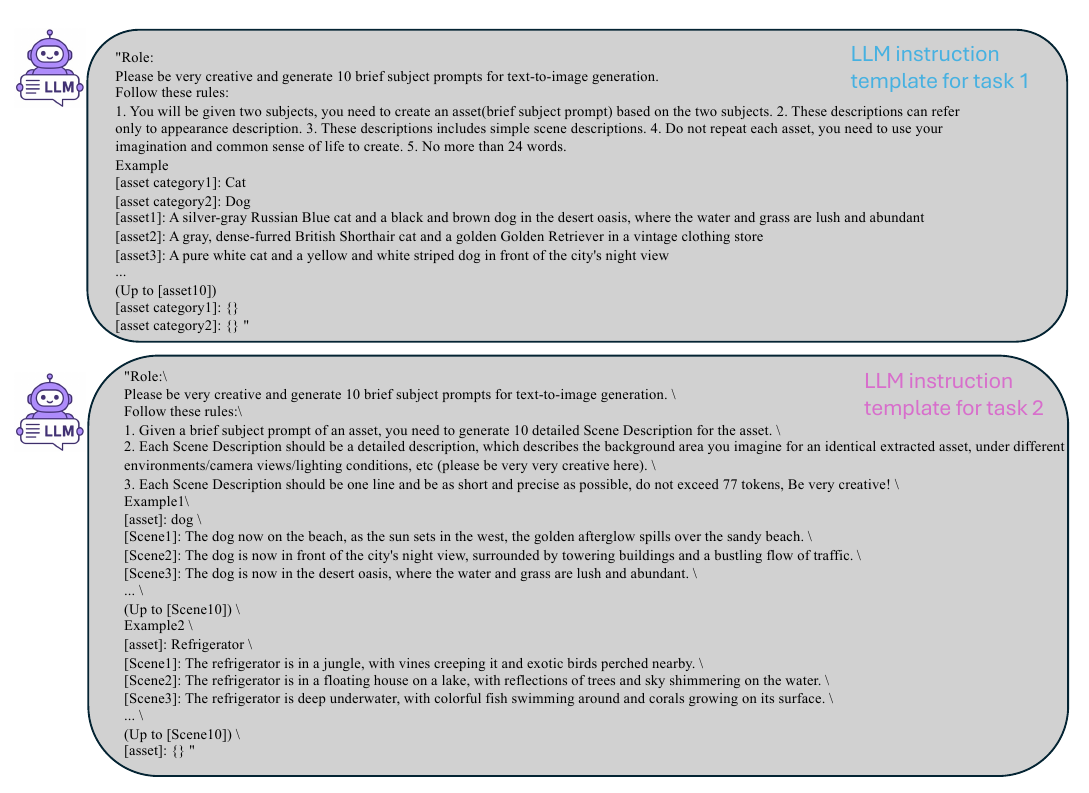}
\caption{Instruction template used for providing to Qwen3 to construct the dataset.}
\label{fig:template}
\end{figure*}

In the construction of the dataset, we use Qwen3-32B~\cite{qwen3} to generate instructions for T2I and single subject personalization, with the prompt template for Qwen3-32B shown in Figure~\ref{fig:template}. We utilize the Text-to-Image model FLUX.1-schnell~\cite{flux} to generate $I_{out}$. During the construction process, since Qwen-Image~\cite{qwen-image} performs better when generating images containing four subjects, we use Qwen-Image~\cite{qwen-image} for generating data involving four subjects. Fig.~\ref{fig:fig-all}-D visualizes the number of samples per category in our dataset. The results show that our dataset exhibits a diverse and relatively balanced category distribution.

\begin{figure*}[h]
\includegraphics[width=1.0\linewidth]{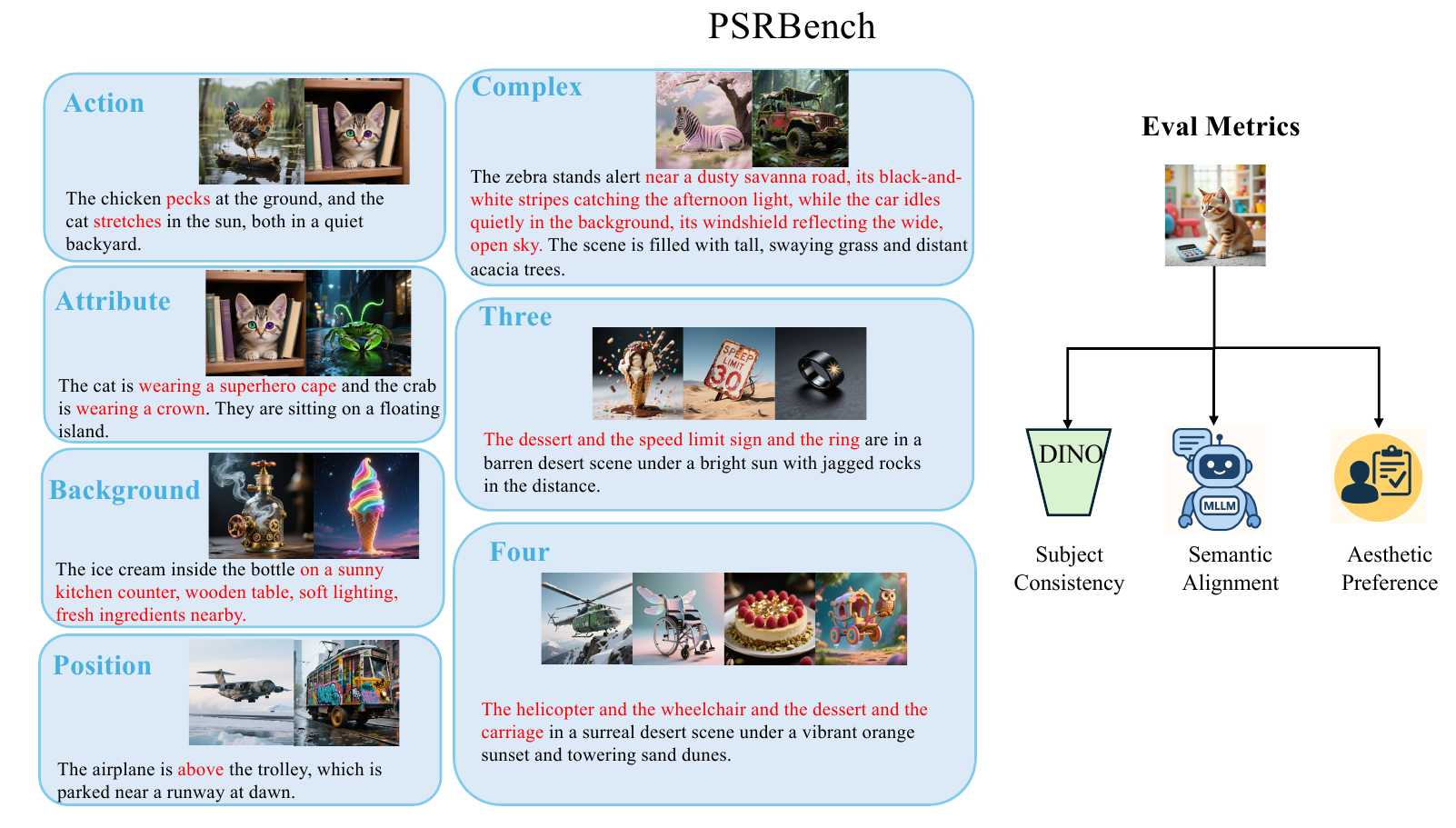}
\caption{Overview of PSRBench, with a case from each subset shown on the left and the three evaluation dimensions for each subset displayed on the right.}
\label{fig:PSRBench}
\end{figure*}

Our benchmark consists of seven subsets, as illustrated in Figure~\ref{fig:PSRBench}, with each subset containing 50 evaluation samples. 
Our input images are generated by Qwen-Image~\cite{qwen-image} and are manually filtered to select images with distinctive subject appearances, ensuring the high quality of the benchmark.
The basic information for each subset is as follows:
\begin{itemize}
\item \textbf{Action:} Each input image in this subset features a different animal, with the prompt explicitly requiring each animal to perform a distinct action, such as ``sit", ``run", and so on. \\
    \item \textbf{Attribute:} This subset also consists of animal images, with the prompt specifying that each animal must possess a unique attribute, such as ``wearing a hat", ``wearing a crown", etc. This is considered a more challenging subset. \\
    \item \textbf{Background:} The input images in this subset can include both objects and animals, with the prompt requiring the images to feature a specific background. \\
    \item \textbf{Position:} The inputs in this subset are arbitrary, with the prompt instructing that the subjects be placed in fixed orientations within the image, such as ``on the right", ``on the left", etc. \\
    \item \textbf{Complex:} The prompts in this subset are more intricate, potentially including detailed backgrounds, specific animal actions, or subjects placed in precise locations within the image. \\
    \item \textbf{Three:} This subset consists of three input subjects, which must be placed within a specific background. \\
    \item \textbf{Four:} This subset contains four input subjects, each requiring a specific background.
\end{itemize}

\begin{figure*}[h]
\includegraphics[width=1.0\linewidth]{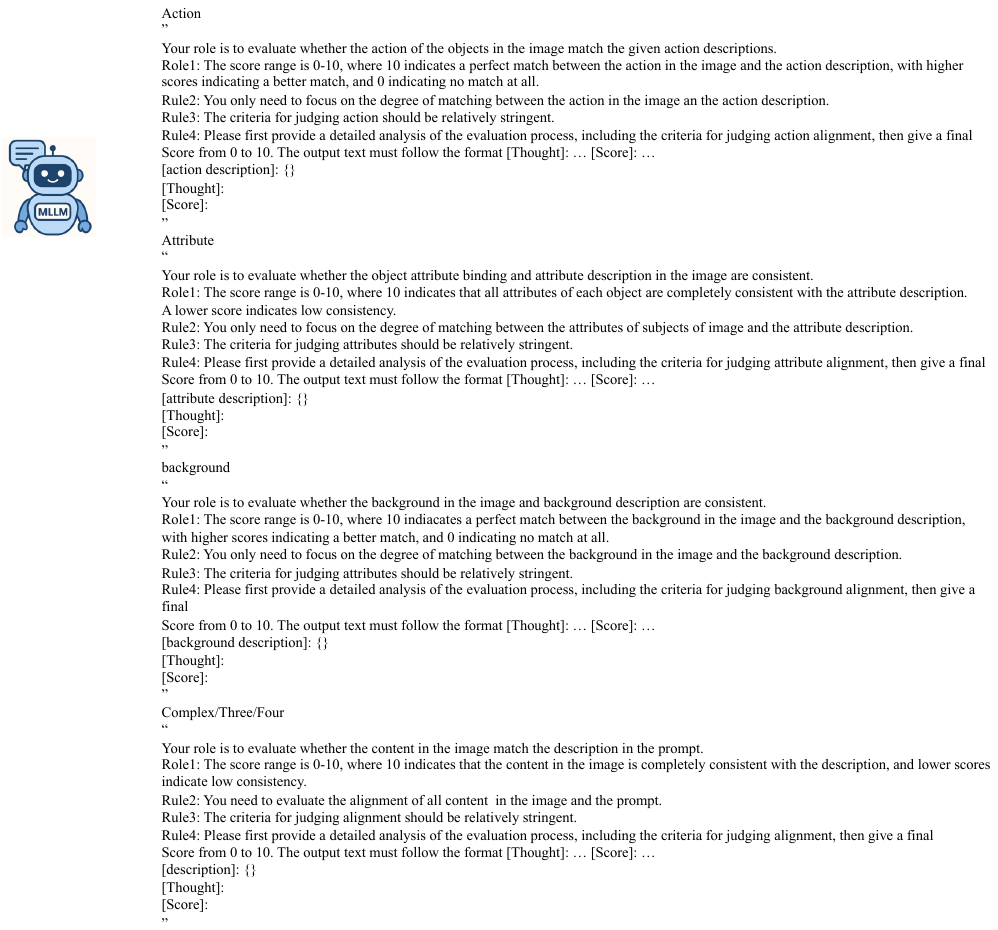}
\caption{Instruction template used for providing to Qwen2.5-VL to evaluate the semantic alignment scores of different subsets.}
\label{fig:eval-template}
\end{figure*}

For evaluation, we consider three complementary dimensions. First, for subject consistency, we detect and crop the corresponding subjects in both the input references and the generated images, compute pairwise similarities, and average the results. Because subject consistency is defined as an averaged similarity score, its value naturally falls within the range of 0 to 1.
For semantic alignment, we employ Qwen2.5-VL-32B-Instruct~\cite{qwen25vl} to assess whether the generated image semantically aligns with the given prompt. The MLLM outputs a score between 0 and 10, where 10 indicates perfect alignment and 0 indicates complete misalignment. We then normalize this score to the range of 0–1. For each specific subset, the corresponding prompt template used for evaluation by the MLLM is shown in Figure~\ref{fig:eval-template}.
For aesthetic preference, we evaluate image quality using HPSv3~\cite{ma2025hpsv3}, an uncertainty-aware ranking model that provides a non-normalized aesthetic score. To make it compatible with other reward components during training, we normalize the score by dividing it by 10.
\section{Data Cleaning}
To obtain high-quality paired data, we further design a stringent data-filtering pipeline. Specifically, we filter samples based on both subject consistency and semantic alignment. Similar to our evaluation protocol, we compute a paired DINO~\cite{dinov2} score to assess the consistency between the input subjects and the generated outputs, and we employ Qwen2.5-VL~\cite{qwen25vl} to evaluate semantic alignment with the prompt. After this filtering stage, we obtain a curated dataset of approximately 350K high-quality multi-subject personalization samples.
\section{Metric Comparison}
In this section, we compare several metrics used to evaluate subject consistency in previous works.

\begin{figure*}[h]
\includegraphics[width=1.0\linewidth]{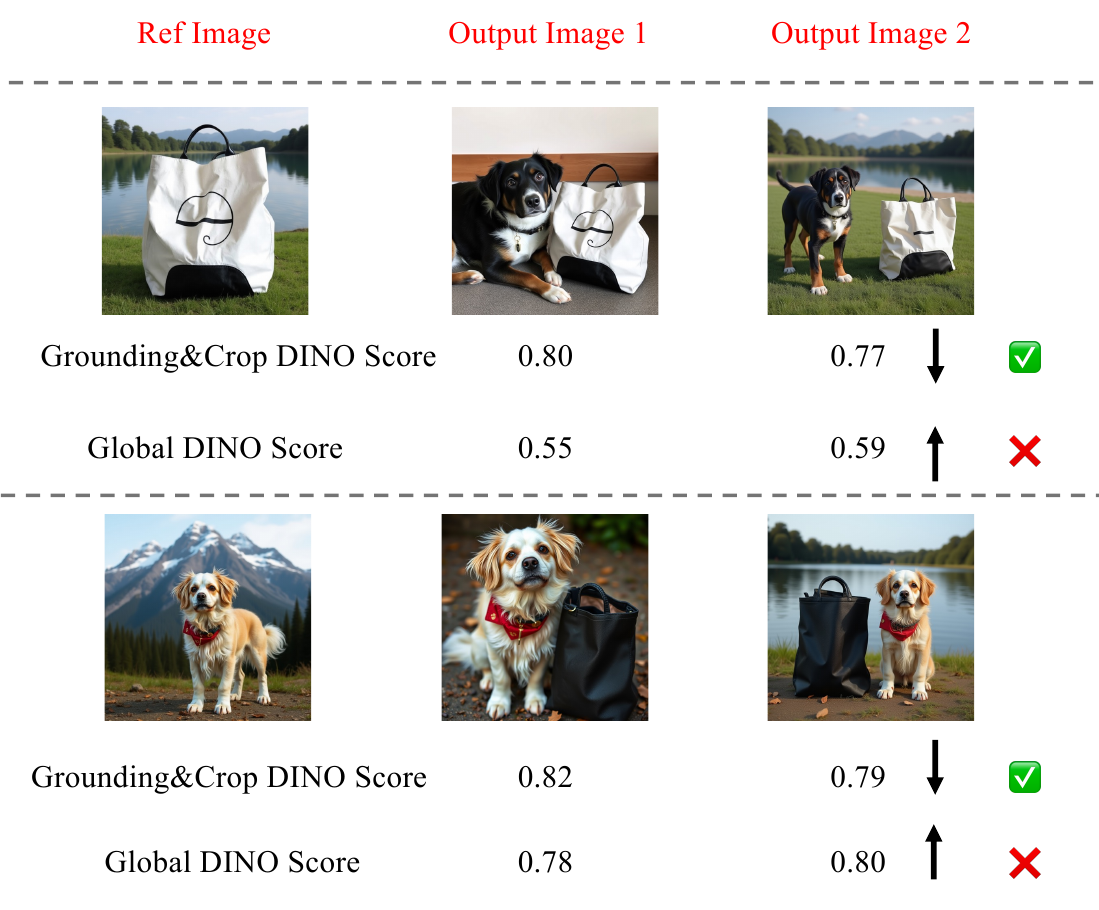}
\caption{Comparison of different metrics.}
\label{fig:metric_1}
\end{figure*}

UNO~\cite{wu2025less} and MIP-Adapter~\cite{MIP-Adapter} assess subject consistency by directly comparing the DINO~\cite{dinov2} score of the input image with that of the output image. However, the output image often involves other subjects and background information, which can significantly affect the evaluation of subject consistency. As shown in Figure~\ref{fig:metric_1}, the results of the traditional Global DINO Score evaluation are presented. It is evident that the bag in `Output Image 1' is more similar to the bag in the `Ref Image'. However, the Global DINO Score assigns a higher score to `Output Image 2'. This is because the background of the reference image and `Output Image 2' are more similar, thus distorting the subject consistency evaluation. In contrast, our method first grounds the subject and then performs cropping, effectively eliminating the irrelevant background information that could interfere with the evaluation. As demonstrated in Figure~\ref{fig:metric_1}, our evaluation metric accurately identifies which image is more similar to the input subject, highlighting the precision of our approach.

\begin{figure*}[h]
\includegraphics[width=1.0\linewidth]{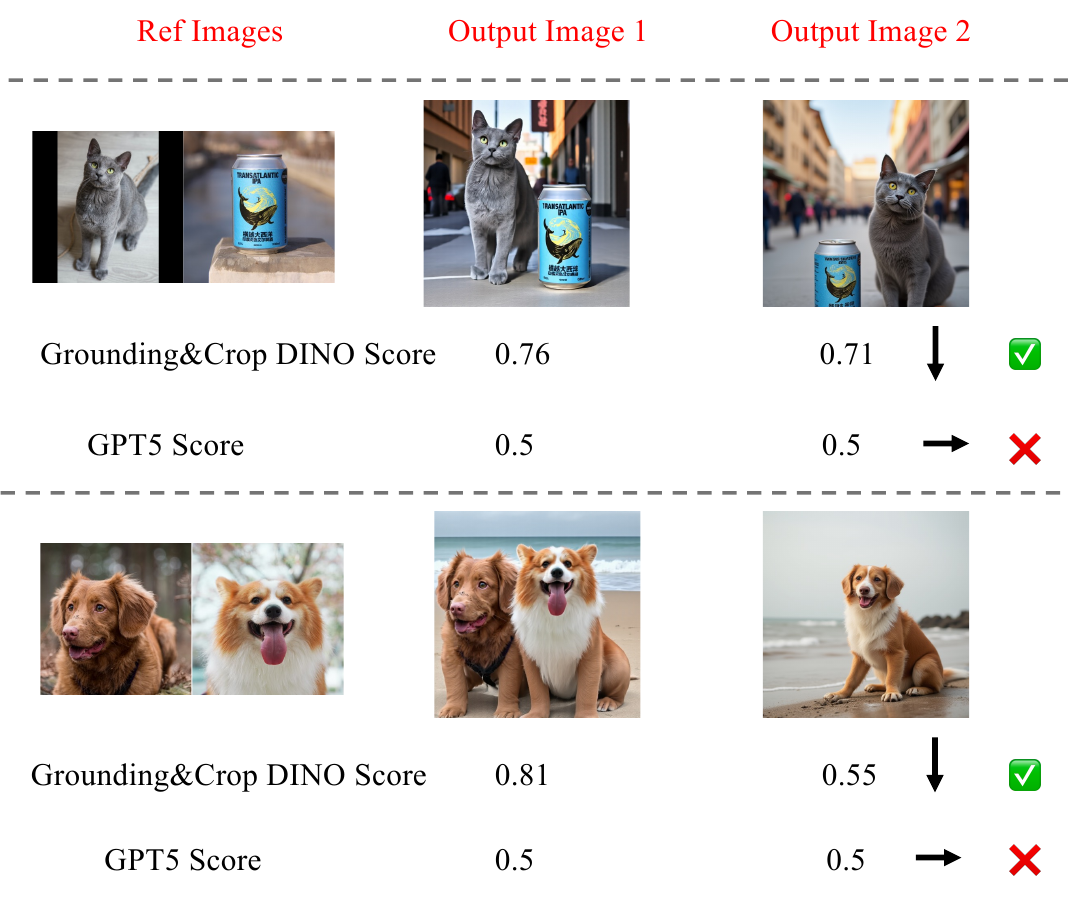}
\caption{Comparison of different metrics.}
\label{fig:metric_mmlm}
\end{figure*}

Some existing methods use multimodal large language models  to evaluate subject consistency. Therefore, we also compare our method with MLLM-based evaluations. We utilize the powerful multimodal large language model GPT-5 and employ the evaluation prompt proposed in OmniContext~\cite{wu2025omnigen2}. As shown in Figure~\ref{fig:metric_mmlm}, even the most advanced GPT-5 model fails to be highly sensitive to changes in the subject’s appearance during subject consistency evaluation, often producing hallucinations.






\clearpage
\clearpage
{
    \small
    \bibliographystyle{ieeenat_fullname}
    \bibliography{main}
}

\end{document}